\documentclass[conference]{IEEEtran}
\IEEEoverridecommandlockouts
\usepackage{cite}
\usepackage{amsmath,amssymb,amsfonts}
\usepackage{algorithmic}
\usepackage{graphicx}
\usepackage{textcomp}
\usepackage{xcolor}

\usepackage{algorithmic}
\usepackage[linesnumbered,ruled]{algorithm2e}
\usepackage{subcaption}
\usepackage{siunitx}
\newcolumntype{P}[1]{>{\centering\arraybackslash}p{#1}}

\def\BibTeX{{\rm B\kern-.05em{\sc i\kern-.025em b}\kern-.08em
    T\kern-.1667em\lower.7ex\hbox{E}\kern-.125emX}}
\begin{document}

\title{Particle Swarm Optimisation for Evolving Deep Neural Networks for Image Classification by Evolving and Stacking Transferable Blocks
}

\author{
	\IEEEauthorblockN{Bin Wang, Bing Xue and Mengjie Zhang}
	\IEEEauthorblockA{School of Engineering and Computer Science\\
		Victoria University of Wellington,
		PO Box 600, Wellington 6140, NEW ZEALAND}
	\IEEEauthorblockA{Emails: \{bin.wang, bing.xue, mengjie.zhang\}@ecs.vuw.ac.nz}
}

\maketitle

\begin{abstract}
Deep Convolutional Neural Networks (CNNs) have been widely used in image classification tasks, but the process of designing CNN architectures is very complex, so Neural Architecture Search (NAS), automatically searching for optimal CNN architectures, has attracted more and more research interests. However, the computational cost of NAS is often too high to be applied to real-life applications. In this paper, an efficient particle swarm optimisation method named EPSOCNN is proposed to evolve CNN architectures inspired by the idea of transfer learning. EPSOCNN successfully reduces the computation cost by minimising the search space to a single block and utilising a small subset of the training set to evaluate CNNs during the evolutionary process. Meanwhile, EPSOCNN also keeps very competitive classification accuracy by stacking the evolved block multiple times to fit the whole training dataset. The proposed EPSOCNN algorithm is evaluated on CIFAR-10 dataset and compared with 13 peer competitors including deep CNNs crafted by hand, learned by reinforcement learning methods and evolved by evolutionary computation approaches. It shows very promising results with regard to the classification accuracy, the number of parameters and the computational cost. Besides, the evolved transferable block from CIFAR-10 is transferred and evaluated on two other datasets --- CIFAR-100 and SVHN. It shows promising results on both of the datasets, which demonstrate the transferability of the evolved block.  All of the experiments have been performed multiple times and Student's t-test is used to compare the proposed method with peer competitors from the statistical point of view.
\end{abstract}


\section{Introduction}

Convolutional Neural Networks (CNNs) have demonstrated the dominance in image classification tasks by continuously increasing the state-of-the-art classification accuracy on various benchmark datasets, from AlexNet \cite{krizhevsky2012imagenet}, VGGNet \cite{simonyan2014very} to very deep CNNs such as ResNet \cite{he2016deep} and DenseNet \cite{huang2017densely}. However, the complex process of designing the above CNN architectures is time-consuming and error-prone, which also requires speciality and expertise in both CNN architectures and the dataset. 

As a result, automatically designing CNN architectures has naturally drawn the research interest. Both reinforcement learning (RL) methods, e.g. \cite{zoph2016neural} and \cite{zoph2018learning}, and evolutionary computation (EC) approaches, such as \cite{wang2018evolving}, \cite{wang2018hybrid}, \cite{xie2017genetic}, and \cite{suganuma2017genetic}, have been used to automatically design CNN architectures in recent years. Although automatically designed CNN architectures have achieved promising results compared to hand-crafted CNNs, it is hard to balance the trade-off between efficiency and effectiveness of automatically designing CNN architectures. For example, \cite{xie2017genetic} worked efficiently by compromising the effectiveness; while \cite{zoph2016neural} and \cite{zoph2018learning} set the state-of-the-art classification accuracy with intimidating computational cost of 22,400 GPU-days for \cite{zoph2016neural} and 2,000 GPU-days for \cite{zoph2018learning}. In this paper, an efficient PSO method will be proposed to evolve CNN architectures without compromising the classification accuracy.

\textbf{Goals:} The overall goal of this paper is to propose an efficient EC-based method to speed up the evolutionary process of automatically designing CNN architectures for image classification without compromising the classification accuracy. The proposed method will be evaluated on the CIFAR-10 dataset and compared with the state-of-the-art methods consisting of hand-crafted CNNs, CNNs found by RL methods, and CNNs obtained by other EC approaches. In addition, the evolved block will be transferred and evaluated on two other datasets --- CIFAR-100 and the Street View House Numbers (SVHN) dataset. The goal will be achieved through the following effort and contributions: 

\begin{figure*}[ht]
	\centering
	\includegraphics[width=0.9\textwidth]{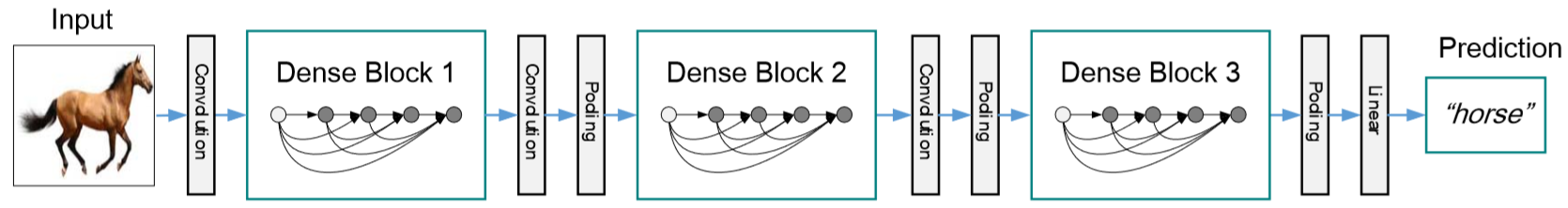}
	\caption{DenseNet architecture (Image taken from \cite{huang2017densely})}
	\label{fig:densenet_architecture}
\end{figure*}

\begin{itemize}
	\item Firstly, the search space is minimised by integrating the existing expertise of hand-crafted CNNs. As the search space of CNNs without any constraints can be infinite, it is not possible to explore the search space both effectively and efficiently. However, by introducing the expertise of hand-crafted CNN architectures, the search space can be considerably reduced. In the proposed method, DenseNet \cite{huang2017densely} is used as the prior expertise to minimise the search space by encoding only the hyper-parameters of one dense block, where the details can be found in Section \ref{SSS:epsocnn_encoding}. The same methodology can be applied to any of the state-of-the-art CNN architectures such as ResNet \cite{he2016deep}.
	\item Secondly, a transferable block is learnt from a small subset of the training dataset. This is inspired by the idea of transfer learning, which is to learn a model on a smaller dataset and transfer the learned model on a larger dataset. In order to mitigate the bias introduced by only using a small subset, Adam optimisation \cite{kingma2014adam} is used to train the CNNs instead of SGD optimisation \cite{sutskever2013importance}. The reasons will be discussed in Section \ref{SSS:epsocnn_fitness}.
	\item Thirdly, an automatic and progressive process of stacking the learned block is proposed to increase the capacity of the final neural network. As the small subset of the dataset may only require a CNN with much less capacity to achieve the best classification accuracy, in order not to compromise the classification accuracy on the whole dataset, the proposed method stacks the learned block multiple times to obtain a CNN with more capacity, which will be depicted in Section \ref{SSS:epsocnn_stack}. 
	\item Lastly, the transferable block learned from one dataset is transferred to two other datasets, which demonstrates the transferability of the learned block. The transferable blocks evolved only once from one dataset will be stacked according to the stacking mentioned in the third effort to form a neural network that can achieve a promising performance without any further NAS process on two other datasets.
\end{itemize}

\section{Background}

\subsection{DenseNet}

A DenseNet \cite{huang2017densely} is composed of several dense blocks illustrated in Fig. \ref{fig:densenet_architecture}. A 1$\times$1 convolutional layer followed by a 2$\times$2 average pooling layer is added to connect dense blocks. The hyper-parameters of dense blocks are dependent on specific image classification tasks, which are the number of layers in the dense block and the \textit{growth rate} of the dense block. The \textit{growth rate} is the number of output feature maps for each convolutional layer in the dense block. The output $x_{l}$ is calculated according to Formula (\ref{eq:DenseNet_ouput}), where $[x_{0}, x_{1}, ..., x_{l-1}]$ refers to the concatenation of the feature maps obtained from layer 0, 1, ..., $l-1$, and $H_{l}$ represents a composite function of three consecutive operations, which are batch normalization (BN) \cite{ioffe2015batch}, a rectified linear unit (ReLU) \cite{nair2010rectified} and $3\times3$ convolution (Conv).

\begin{equation}\label{eq:DenseNet_ouput}
x_{l} = H_{l}([x_{0}, x_{1}, ..., x_{l-1}])
\end{equation}

\subsection{Particle Swarm Optimisation}

Particle Swarm Optimization (PSO) \cite{eberhart1995new} is a population-based EC algorithm, which can be used for solving optimization problems lacking of domain knowledge. The population is constituted of a number of particles. Each of them represents a candidate. It searches for the best solution by updating velocity and particle vector according to Equations (\ref{eq:UpdateV}) and (\ref{eq:UpdateX}), respectively, where $v_{id}$ represents the velocity of the particle $i$ in the $d$th dimension, $x_{id}$ represents the position of particle $i$ in the $d$th dimension, $P_{id}$ and $P_{gd}$ are the local best and the global best in the $d$th dimension, $r_{1}, r_{2}$ are random numbers between 0 and 1, $w, c_{1}$ and $c_{2}$ are inertia weight and acceleration coefficients for exploitation and acceleration coefficient for exploration, respectively. Since the encoded vector in the proposed method is fixed-length and consists of decimal values, and PSO is effective to search for the optimal solution in a fixed-length search space of decimal values, the proposed method will use PSO as the search algorithm. 

\begin{equation}\label{eq:UpdateV}
v_{id}(t+1) = w * v_{id}(t) + c_{1} * r_{1} * (P_{id} - x_{id}(t)) + \\
c_{2} * r_{2} * (P_{gd} - x_{id}(t))
\end{equation}

\begin{equation}\label{eq:UpdateX}
x_{id}(t+1) = x_{id}(t) + v_{id}(t+1)
\end{equation}

\subsection{Related Work}

One of the research works of employing EC to evolve CNNs is GeNet \cite{xie2017genetic}. GeNet proposed an encoding strategy of using a fixed-length binary string to represent CNN architectures, with which the standard Genetic Algorithm can be easily applied.  It can be noticed that GeNet only consumes 20 GPU-days to evolve CNN architectures, but it has compromised the classification accuracy, which we believe may be because of not reaching the global optima or some good local optima near the global optima due to the huge search space.

Another successful EC method is AmoebaNet \cite{real2018regularized}, which performs EC search in the same search space as that of NAS \cite{zoph2016neural}. It set up the state-of-the-art classification accuracy on several image classification tasks, which was the first time for utilising an EC method to produce the state-of-the-art image classifier; however, the computational cost has gone much higher than that of GeNet, which rose significantly to 3,150 GPU-days approximately. 

In summary, it can be seen that it is hard to balance the trade-off between efficiency and effectiveness of automatically designing CNN architectures as GeNet worked efficiently by compromising the effectiveness and AmoebaNet set the state-of-the-art classification accuracy with very high computational cost. In this paper, an efficient PSO method will be proposed to evolve CNN architectures without compromising the classification accuracy. 

\section{The Proposed Method}

The proposed method will achieve the goal of efficiently evolving deep CNNs without compromising the classification accuracy through the following aspects. Firstly, the encoding strategy will minimise the search space by leveraging the expertise of the state-of-the-art CNN architecture. Secondly, a transferable block will be learned from a small subset of training dataset instead of the whole training dataset, which will considerably reduce the time consumed by fitness evaluation. Thirdly, the evolved block will be stacked multiple times to improve the classification accuracy. In the end, the transferable block learned from the CIFAR-10 dataset will be transferred to the CIFAR-100 and SVHN datasets.

\subsection{Overall Framework}

\begin{figure}
	\centering
	\includegraphics[width=\linewidth]{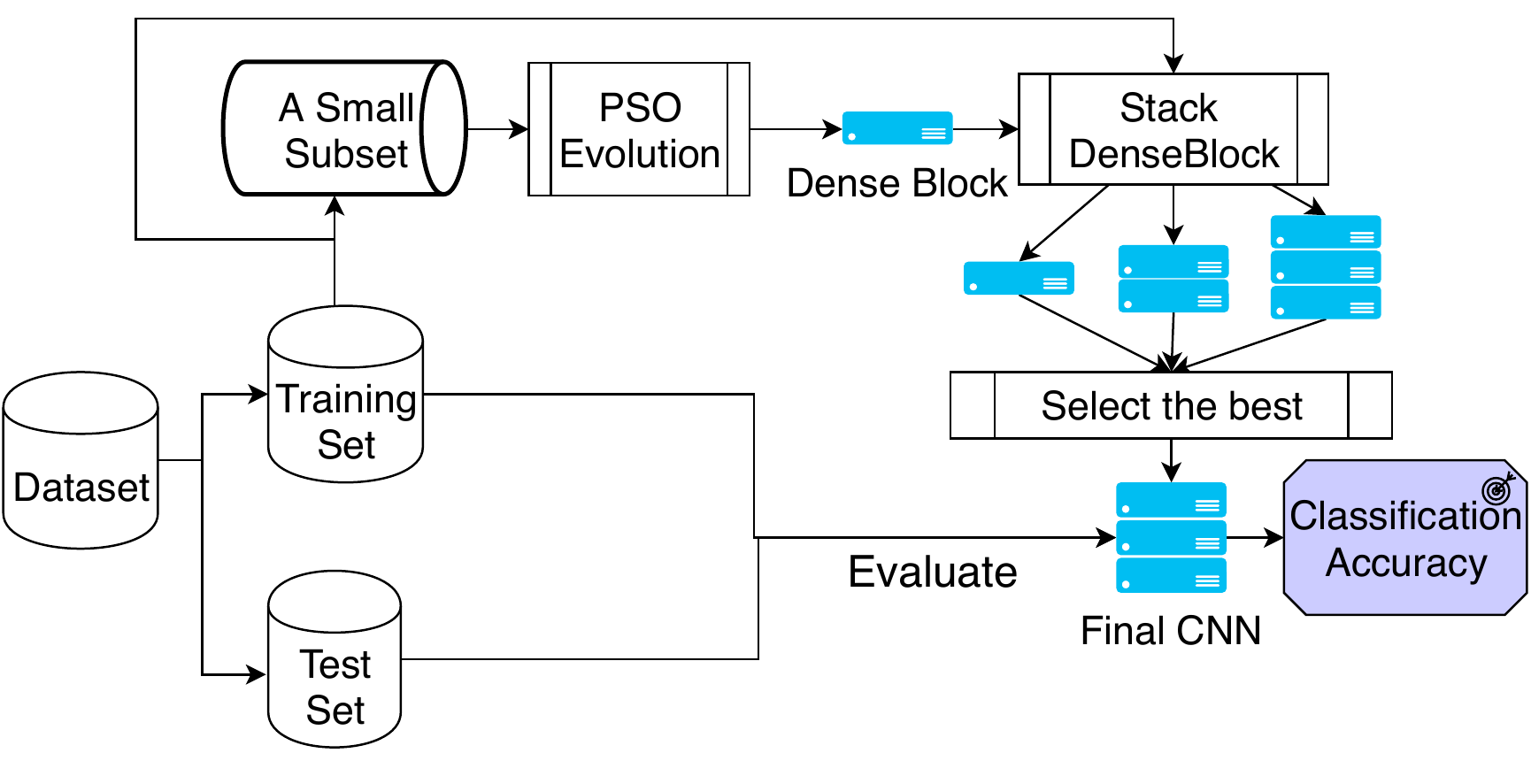}
	\caption{The overall framework of the proposed method}
	\label{fig:epsocnn_framework}
\end{figure}

The overall framework of the proposed method is illustrated in Fig. \ref{fig:epsocnn_framework}. Firstly, the dataset is split into a training set and a test set, and then a small subset of the training set is randomly sampled from the training set, which will be passed to the PSO evolutionary process. Furthermore, the small subset is used during the PSO evolutionary process. The primary reason of using a small subset of the training set for fitness evaluation is to reduce the computational cost because given a CNN architecture, it takes less time to train it and requires less memory when the dataset is smaller, which will be further discussed in Section \ref{SSS:epsocnn_fitness}. Instead of evolving the whole network architecture, the PSO is only utilised to evolve the optimal Dense Block on the small subset. The details of the intuition will be analysed in Section \ref{SSS:epsocnn_encoding}. Next, the proposed method stacks the evolved Dense Block various number of times to produce a set of CNN architectures, and the best CNN architecture is then selected to be the final evolved CNN architecture, where the details will be written in Section \ref{SSS:epsocnn_stack}. Finally, the classification accuracy of the final CNN architecture will be reported. 

\subsection{Encoding Strategy to Minimise The Search Space}\label{SSS:epsocnn_encoding}

As the proposed method concentrates on the efficiency, the encoding strategy endeavours to minimise the search space. Inspired by the idea of learning the CNN architecture from smaller datasets and transferring the learned architecture to larger datasets \cite{zoph2018learning}, instead of evolving multiple blocks as \cite{wang2019hybrid}, the proposed method only evolves a single dense block. By doing the simplification of the encoding strategy, the search space has been significantly reduced, which has turned a complex search problem to a simple search task. In the research of learning transferable architectures \cite{zoph2018learning}, it has proved that the learned block was transferable from CIFAR-10 dataset to ImageNet dataset. Inferring from that, the block learned from the small subset of the training set is more likely to be transferable to the whole training dataset, which will be further tested by the experimental results. Therefore, the final encoded vector only consists of two dimensions, which are the \textit{growth rate} and the \textit{number of layers}. 

\subsection{Fitness Evaluation of Transferable Blocks}\label{SSS:epsocnn_fitness}

Fitness evaluation of PSO, whose pseudo-code is shown in Algorithm \ref{alg:epsocnn_fitness}, is crucial to the computational cost of the proposed method. Since the essential target of the proposed method is to boost the efficiency of evolving deep CNNs, two methods are utilised to reduce the computational cost. Firstly, only a small subset of training dataset is used to train and evaluate the particle, which holds the hyper-parameters of the dense block. Therefore, both the memory required to train the CNNs and the training time will be reduced. Furthermore, an adaptive training algorithm referred as Adam optimisation \cite{kingma2014adam} is adopted to train the CNNs during the fitness evaluation, whose ability of fast convergence has been proved, so the fitness evaluation using Adam optimisation is faster than those relying on Stochastic Gradient Descent (SGD) \cite{sutskever2013importance}. 
By combining the strategies of cutting down the training data for fitness evaluation and adopting the adaptive training algorithm, the proposed method is expected to reduce the computational cost. 

\begin{algorithm}[t]
	\caption{Fitness Evaluation}
	\label{alg:epsocnn_fitness}
	\begin{algorithmic}[1]
		\renewcommand{\algorithmicrequire}{\textbf{Input:}}
		\renewcommand{\algorithmicensure}{\textbf{Output:}}
		\newcommand{\algorithmicbreak}{\textbf{break}}
		\newcommand{\BREAK}{\STATE \algorithmicbreak}
		\REQUIRE number of layers $nol$, growth rate $gr$, a small subset of training set $dt$;
		\STATE $acc_{best}, epoch_{best}, epoch, acc\leftarrow$ 0, 0, 0, 0;
		\STATE $dt_{train}, dt_{test}\leftarrow$ Randomly split $dt$ into 80\% as training part $dt_{train}$ and 20\% test part $dt_{test}$;
		\STATE $block\leftarrow$ Build the dense block according to $nol$ and $gr$;
		\WHILE{$acc>=acc_{best}$ \textbf{or} $epoch-epoch_{best}<5$}
		\STATE Apply Adam optimisation \cite{kingma2014adam} to train $block$ on $dt_{train}$ for one epoch;
		\STATE $acc\leftarrow$ Evaluate $block$ on $dt_{test}$;
		\IF{$acc > acc_{best}$}
		\STATE $acc_{best}, epoch_{best}\leftarrow$ $acc, epoch$;
		\ENDIF
		\STATE $epoch\leftarrow$ $epoch + 1$;
		\ENDWHILE
		\RETURN $acc_{best}$
	\end{algorithmic}
\end{algorithm}

Apart from the perspective of improving the efficiency, the proposed fitness evaluation arguably tends to select the best transferable block. In the research of learning transferable block \cite{zoph2018learning}, it has been proved that the learned block from a smaller dataset is transferable to a larger dataset even though the two datasets are quite different both in terms of the number of classes and the image resolutions. By following the same idea, the small subset of training data is sampled to only learn a single block instead of learning the final CNN architecture. Since the similarity between the sampled small subset and the target dataset is closer than the similarity between two different datasets, the theory of learning transferable block should also apply to the proposed method, which is expected to produce a good classification accuracy when stacking the learned block multiple times for the whole dataset. In addition, Adam optimisation is designed to adapt the learning rate according to the specific CNN during the training process; while, when using other non-adaptive optimisation such as SGD, they need to be fine-tuned for a specific CNN on a particular dataset. Reversely, if non-adaptive optimisation is adopted with a specific set of settings, the specific CNN or similar CNNs are more likely to be chosen \cite{wang2019evolving}. Therefore, the fitness value obtained by using Adam optimisation can represent the quality of the block. 

\subsection{Evolving Dense Block By PSO}\label{SSS:epsocnn_evolve}

\begin{algorithm}[ht]
	\caption{Evolving dense block by PSO}
	\label{alg:epsocnn_pso}
	\begin{algorithmic}[1]
		\renewcommand{\algorithmicrequire}{\textbf{Input:}}
		\renewcommand{\algorithmicensure}{\textbf{Output:}}
		\newcommand{\algorithmicbreak}{\textbf{break}}
		\newcommand{\BREAK}{\STATE \algorithmicbreak}
		\REQUIRE population size $pop_{size}$, generations $gen$;
		\STATE $pop\leftarrow$ Random initialise the particles until the number of particles reaches $pop_{size}$;
		\STATE $g_{best}, i\leftarrow$ Empty, 0; 
		\WHILE{$i<gen$}
		\FOR{particle $p$ in $pop$}
		\STATE $p\leftarrow$ Apply standard PSO operations to update the position of $p$;
		\STATE $fitness\leftarrow$ Use fitness evaluation in Algorithm \ref{alg:epsocnn_fitness} to calculate the fitness for $p$;
		\STATE Update the fitness of $p$ by $fitness$;
		\IF{$fitness>$ \textit{the fitness of the personal best}}
		\STATE Update the person best of $p$ with the current $p$;
		\ENDIF
		\ENDFOR
		\STATE $g_{best}\leftarrow$ Update with the best particle among the current $g_{best}$ and $pop$;
		\STATE $i\leftarrow$ $i+1$;
		\ENDWHILE
		\RETURN $g_{best}$;
	\end{algorithmic}
\end{algorithm}

After simplifying the encoded vector by the proposed encoding strategy, standard PSO can be applied to solve the optimisation tasks, which is depicted in Algorithm \ref{alg:epsocnn_pso}. 
There are still a couple of points that need to be carefully designed. Firstly, due to the memory limit, the number of layers and the growth rate are required to be constrained into a range. On the other hand, the lower bound of the these two hyper-parameters also need to be restricted as too few layers in a block will not provide enough capacity for the model and a too small growth rate will not capture enough feature information. Therefore, two hyper-parameters of the proposed method need to be defined before running it, which are the range of the number of layers $\alpha_{l}$ and the range of growth rate $\alpha_{g}$; however, these two hyper-parameters are easy to choose as the default values of $\alpha_{l}$ and $\alpha_{g}$ will suit most of the image classification tasks. When initialising the particles, the number of layers and the growth rate are randomly sampled within the range of $\alpha_{l}$ and $\alpha_{g}$, respectively. Secondly, during the evolutionary process, the values of the hyper-parameters may fall out of the range, so the proposed PSO method needs to rectify these values by setting them to either the upper bound or the lower bound of the range. Furthermore, as some of the dense block may cause out of memory issue, the proposed PSO method captures this error and set the fitness of particles to zero in order to eliminate them. 

\subsection{Stacking and Selecting Best CNN}\label{SSS:epsocnn_stack}

After the dense block is obtained by PSO as shown in Algorithm \ref{alg:epsocnn_pso}, the dense block is stacked multiple times to produce the candidates for the final model by progressively stacking the dense block, i.e. the number of dense blocks starts from 1 and increases by 1 each time. The whole training set is used to train and evaluate the stacked candidate, and Adam optimisation is chosen to train the stacked CNNs with the same reasons described in Section \ref{SSS:epsocnn_fitness}. Once the classification accuracy of the stacked CNN candidate does not increase or the stacked CNN requires more memory than the hardware resource, the stacking process stops and the best candidate so far is selected as the final solution. 

\begin{algorithm}[ht]
	\caption{Progressively stack and select the best candidate}
	\label{alg:epsocnn_stack}
	\begin{algorithmic}[1]
		\renewcommand{\algorithmicrequire}{\textbf{Input:}}
		\renewcommand{\algorithmicensure}{\textbf{Output:}}
		\newcommand{\algorithmicbreak}{\textbf{break}}
		\newcommand{\BREAK}{\STATE \algorithmicbreak}
		\REQUIRE Evolved block $b$, the whole training set $dt$;
		\STATE $dt_{train}, dt_{test}\leftarrow$ Randomly split $dt$ into 80\% as training part $dt_{train}$ and 20\% test part $dt_{test}$;
		\STATE $c_{best}, i, acc\leftarrow$ Empty best candidate, 0 as stacking times, 0 as the accuracy of the stacked CNN; 
		\WHILE{True}
		\STATE $i\leftarrow$ $i+1$
		\STATE $c\leftarrow$ Stack $b$ for $i$ times to generate a candidate;
		\STATE Use Adam optimisation to train $c$ on $dt_{train}$ with the same stopping criteria as that of fitness evaluation in Section \ref{SSS:epsocnn_fitness};
		\STATE $acc\leftarrow$ Evaluate the trained $c$ on $dt_{test}$;
		\IF{$acc>$ \textit{accuracy of} $c_{best}$}
		\STATE $c_{best}\leftarrow$ $c$;
		\ELSE
		\BREAK
		\ENDIF
		\ENDWHILE
		\RETURN $c_{best}$
	\end{algorithmic}
\end{algorithm}

\subsection{Transferring the Evolved Block}\label{SSS:epsocnn_transfer}

To validate the transferability of the evolved block, the proposed method will learn the block on a small subset of one dataset as the transferable block as shown in Section \ref{SSS:epsocnn_evolve}. By following the stacking method in Section \ref{SSS:epsocnn_stack}, the transferable block will be stacked and selected for other datasets without re-evolving the block. By applying transfer learning of the transferable block, it will only evolve the transferable block once even for multiple image classification tasks. Given the evolutionary process consumes most of the computational cost of the proposed method, the proposed method will improve the efficiency of NAS especially when multiple image classification tasks need to be solved. 

\section{Experiment Design}

As EC is a meta-heuristic approach, it requires multiple runs of one experiment and statistical analysis to draw a conclusion. In our experiments, 10 runs are performed due to the limited computational resources and Student's t-test is performed when comparing the proposed method with peer competitors. 

\subsection{Benchmark Datasets and Peer Competitors}\label{SSS:epsocnn_dataset}

CIFAR-10 \cite{krizhevsky2009learning} is used as the benchmark dataset to evaluate the proposed method. It is composed of 60,000 colour images in 10 classes, with 6000 images in each class. The whole dataset is split into training images of 50,000 and test images of 10,000. CIFAR-10 is chosen because it has been widely used to evaluate image classifiers, especially for automatically designing CNN architectures. The results of the peer competitors' performance can be easily collected for comparisons. 

The peer competitors of the CIFAR-10 dataset are selected based on their performance and the relevance to the proposed methods. Firstly, two state-of-the-art hand-crafted CNN architectures will be compared, which are ResNet \cite{he2016deep} and DenseNet \cite{huang2017densely}. Secondly, several state-of-the-art CNN architectures designed by automatic approaches are chosen, which are categorised into two types. The first type utilises RL methods, which includes EAS \cite{cai2018efficient}, NASNet \cite{zoph2018learning}, NASH \cite{elsken2017simple} and NAS \cite{zoph2016neural}. The other type employs EC algorithms to automatically evolve CNN architectures, which are AmoebaNet \cite{real2018regularized}, Hier. repr-n \cite{liu2017hierarchical}, CGP-CNN \cite{suganuma2017genetic}, DENSER \cite{assunccao2018evolving}, GeNet \cite{xie2017genetic}, CoDeapNEAT \cite{miikkulainen2019evolving} and LS-Evolution \cite{real2017large}. 

In order to demonstrate the transferability of the learned block, two other datasets are used to evaluate the transferable blocks. The first dataset is the CIFAR-100 dataset \cite{krizhevsky2009learning}. It consists of the same number of images with the same resolution as the CIFAR-10 dataset, so the two domains of the CIFAR-10 and CIFAR-100 datasets are very similar. However, the number of classes is 100, which makes the classification task more difficult. Therefore, CIFAR-100 is a good benchmark dataset to evaluate the transferable block evolved from a simpler dataset. The SVHN dataset \cite{netzer2011reading} is used as another dataset to evaluate the transferable block. It includes 73,257 images in the training set, 26,032 images in the test set, and 531,131 images for additional training. The two domains of the SVHN and CIFAR-10 datasets are disparate, but the number of classes is the same, so SVHN is suitable to evaluate how the transferable block performs on a different domain with the similar classification difficulty. 

The performance results from six peer competitors on CIFAR-100 and SVHN are collected and used in the comparisons to show the transferability of the transferable block. These peer competitors are recent CNN architectures manually designed by experts, which are FractalNet \cite{larsson2016fractalnet}, Deeply Supervised Net \cite{lee2015deeply}, Network in Network \cite{lin2013network}, Wide ResNet \cite{zagoruyko2016wide}, ResNet \cite{he2016deep} and DenseNet \cite{huang2017densely}.

\subsection{Parameter Settings}\label{SSS:epsocnn_parameter}

The parameters of the proposed method are listed in Table \ref{table:epsocnn_parameters}. The two parameters of the proposed method, the range of number of layers $\alpha_{l}$ and the range of growth rate $\alpha_{g}$, are defined based on the GPU card (GeForce GTX 1080) used to run the experiment. The parameters of PSO is set according to \cite{van2006study}. Based on the computational resource and time limit, 20 has been chosen for both the population size and the number of generations. In terms of the Adam Optimisation used in both the fitness evaluation in Section \ref{SSS:epsocnn_fitness} and the stacking process in Section \ref{SSS:epsocnn_stack}, the default settings described in the study \cite{kingma2014adam} are utilised. 

\begin{table}[t]
	\renewcommand{\arraystretch}{1.3}
	\caption{EPSOCNN parameter settings}
	\label{table:epsocnn_parameters}
	\centering
	\begin{tabular}{|c|c|}
		\hline
		Parameter & Value\\
		\hline
		\multicolumn{2}{|c|}{\textbf{EPSOCNN default hyper-parameters}} \\
		\hline
		range of number of layers $\alpha_{l}$ & [6, 32]\\
		\hline
		range growth rate $\alpha_{g}$ & [12, 32]\\
		\hline
		\multicolumn{2}{|c|}{\textbf{PSO parameters}} \\
		\hline
		inertia weight $w$ & 0.7298\\
		\hline
		acceleration coefficient $c1$ & 1.49618\\
		\hline
		acceleration coefficient $c2$ & 1.49618\\
		\hline
		population size & 20\\
		\hline
		number of generations & 20\\
		\hline
	\end{tabular}
\end{table}

In order to evaluate the final evolved CNN architecture and perform a fair comparison, the same training strategy adopted by most of the peer competitors is adopted. The evolved CNN is trained for 300 epochs and 40 epochs for the CIFAR and SVHN datasets, respectively, by following DenseNet \cite{huang2017densely}. The initial learning rate is set to 0.1, which is divided by 10 at 50\% and 75\% of the total number of training epochs. The weight decay and Nesterov momentum \cite{sutskever2013importance} are set to \num{1e-4} and 0.9 without dampening, respectively. A standard data augmentation strategy \cite{huang2017densely} and weight initialisation method \cite{he2015delving} are used.

\section{Result Analysis}

\subsection{Performance Comparisons}\label{SSS:epsocnn_accuracy}

\subsubsection{Performance Comparisons on CIFAR-10}

\begin{table}[t]
	\renewcommand{\arraystretch}{1.3}
	\caption{Performance comparison with peer competitors on CIFAR-10}
	\label{table:epsocnn_performance}
	\centering
	\begin{tabular}{|P{0.2\linewidth}|P{0.2\linewidth}|P{0.2\linewidth}|P{0.2\linewidth}|}
		\hline
		Method & CIFAR-10 (Error rate\%) & Number of Parameters & Computational Cost\\
		\hline
		ResNet-110 \cite{he2016deep} & 6.43 & \textbf{1.7M} & --\\
		\hline
		DenseNet(k = 40) \cite{huang2017densely} & 3.74 & 27.2M & --\\
		\hline
		\hline
		EAS \cite{cai2018efficient} & 4.23 & 23.4M & $<$10 GPU-days\\
		\hline
		NASNet-A (7 @ 2304) \cite{zoph2018learning} & \textbf{2.97} & 27.6M & 2,000 GPU-days\\
		\hline
		NASH (ensemble across runs) \cite{elsken2017simple} & 4.40 & 88M & 4 GPU-days\\
		\hline
		NAS v3 max pooling \cite{zoph2016neural} & 4.47 & 7.1M & 22,400 GPU-days\\
		\hline
		\hline
		AmoebaNet-B (6,128) \cite{real2018regularized} & \textbf{2.98} & 34.9M & 3150 GPU-days\\
		\hline
		Hier. repr-n, evolution (7000 samples) \cite{liu2017hierarchical} & 3.75 & -- & 300 GPU-days\\
		\hline
		CGP-CNN(ResSet) \cite{suganuma2017genetic} & 5.98 & \textbf{1.68M} & 29.8 GPU-days\\
		\hline
		DENSER \cite{assunccao2018evolving} & 5.87 & 10.81M & --\\
		\hline 
		GeNet from WRN \cite{xie2017genetic} & 5.39 & -- & 100 GPU-days\\
		\hline
		CoDeapNEAT \cite{miikkulainen2019evolving} & 7.3 & -- & --\\
		\hline
		LS-Evolution \cite{real2017large} & 4.4 & 40.4M & $>$2,730 GPU-days\\
		\hline
		\hline
		\textbf{EPSOCNN (Best classification accuracy)} & 3.58 & 6.74M & $<$4 GPU-days\\
		\hline
		\textbf{EPSOCNN (10 runs)} & 3.74$\pm$0.0154 & 4.79M$\pm$1.5363M & $<$4 GPU-days\\
		\hline
	\end{tabular}
\end{table}

The classification error rate, number of parameters and the computational cost of searching for the CNN architecture are listed in Table
\ref{table:epsocnn_performance}. The best results and the mean and standard deviation values from 10 runs of the proposed method are reported and compared with all the peer competitors in three aspects. Firstly, the best classification accuracy of the proposed method is the third, but when comparing the number of parameters between the proposed method and the other two peer competitors having better classification accuracy, the model size is much smaller, which is only less than one-third of either of the other two models. By applying Student's t-test on the 10 runs' results of the proposed method and other peer competitors, the proposed method does not outperform DenseNet(k = 40) and Hier. repr-n, evolution (7000 samples) because the differences are not statistically significant. Secondly, only two of the peer competitors have fewer parameters than the proposed method that achieved the best classification accuracy; however, the error rate of the other two is more than 2\%  larger than that of the proposed method. When performing the statistical analysis on the 10 runs' results, the differences between the proposed method and all others are significant. None of them is better than the proposed method in both the classification accuracy and the number of parameters. Lastly but not least, all of the CNN architectures from the 10 runs are achieved within 4 GPU-days. In comparison with other peer competitors, the computational cost taken to automatically design the CNN architecture of the proposed method is the smallest among all of the peer competitors. Therefore, it can be concluded that the proposed method has achieved very promising and competitive result both in terms of classification accuracy and the number of parameters, and it is the most efficient approach among all of the peer competitors.  

\subsubsection{Transferability on CIFAR-100 and SVHN}

\begin{table}[ht]
	\renewcommand{\arraystretch}{1.3}
	\caption{Error rate comparison with peer competitors on CIFAR-100 and SVHN}
	\label{table:epsocnn_performance_tl}
	\centering
	\begin{tabular}{|P{0.2\linewidth}|P{0.2\linewidth}|P{0.2\linewidth}|}
		\hline
		Method & CIFAR-100 & SVHN\\
		\hline
		\hline
		Network in Network \cite{lin2013network} & 35.68 & 2.35 \\
		\hline
		Deeply Supervised Net \cite{lee2015deeply} & 34.57 & 1.92 \\
		\hline
		FractalNet \cite{larsson2016fractalnet} & 23.30 & 2.01 \\
		\hline
		Wide ResNet \cite{zagoruyko2016wide} & 22.07 & 1.85 \\
		\hline
		ResNet \cite{huang2016deep} & 27.22 & 2.01 \\
		\hline
		DenseNet(k=12) \cite{huang2017densely} & 20.20 & 1.67 \\
		\hline
		\hline
		\textbf{EPSOCNN (Best)} & 18.56 & 1.84 \\
		\hline
		\textbf{EPSOCNN (10 runs)} & 19.05$\pm$0.1874 & 1.89$\pm$0.0387 \\
		\hline
	\end{tabular}
\end{table}

To demonstrate the transferability of the learned block in Section \ref{SSS:epsocnn_transfer}, the learned block on the CIFAR-10 dataset is transferred to two different domains --- the CIFAR-100 and SVHN datasets. The classification error rates are listed in Table \ref{table:epsocnn_performance_tl}. Firstly, on CIFAR-100, the proposed method achieves the best classification among all the peer competitors with almost 2\% more accuracy than DenseNet(k=12), which is the second place. By performing Student's t-test on the results from 10 runs of the proposed method with others, the proposed method outperforms all others with statistical significances. It demonstrates that the transferable block can be transferred to a similar domain with more complexities. Furthermore, on SVHN, the best classification accuracy of the proposed method is the second best among all peer competitors. According to the statistical analysis of Student's t-test, the proposed method ranks the third best, significantly better than four other peer competitors. Overall, the transferable block has shown promising results on the CIFAR-100 and SVHN datasets. 

\subsection{Convergence Analysis}\label{SSS:epsocnn_convergence}

\begin{figure*}[ht]
	\centering
	\begin{subfigure}[b]{0.48\textwidth}   
		\centering 
		\includegraphics[width=\textwidth]{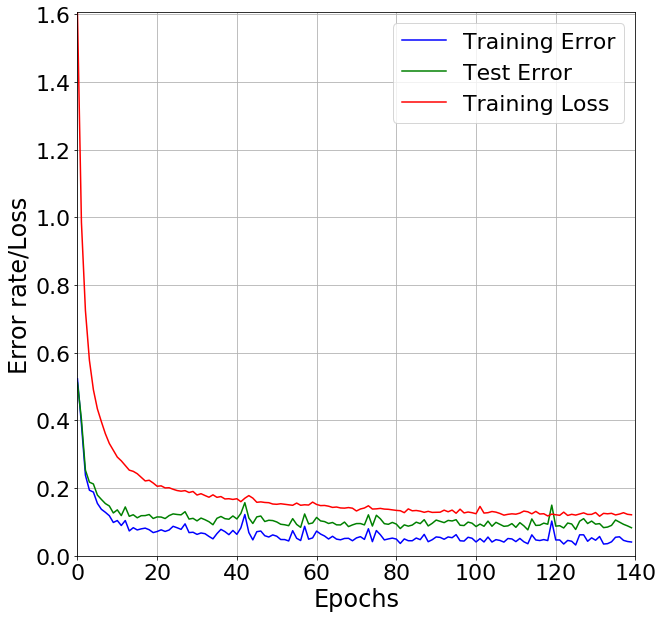}
	\end{subfigure}
	\begin{subfigure}[b]{0.48\textwidth}   
		\centering 
		\includegraphics[width=\textwidth]{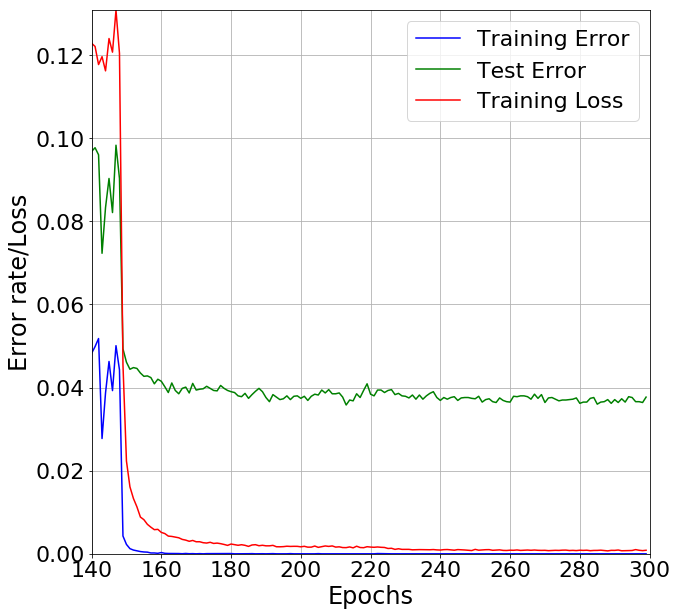}
	\end{subfigure}
	\caption{The training process of the final evolved CNN on CIFAR-10. X-axis: epochs; Y-axis: Error rate/Loss.}
	\label{fig:epsocnn_train}
\end{figure*}

\begin{figure}[ht]
	\centering
	\includegraphics[width=.96\linewidth]{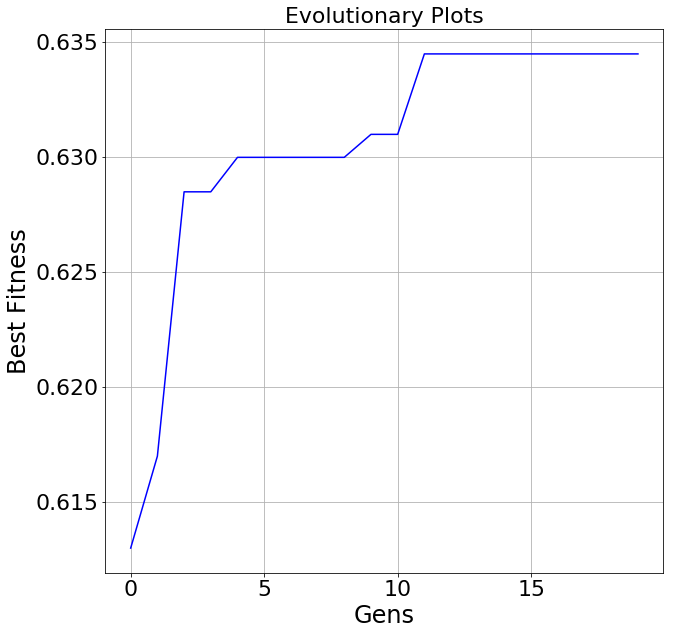}
	\caption{Evolutionary plots. X-axis: generations; Y-axis: best fitness value.}
	\label{fig:epsocnn_trajectory}
\end{figure}



The convergence of the proposed method is shown in Fig. \ref{fig:epsocnn_trajectory}. At the beginning of the evolutionary process, the fitness of the global best particle grows fast until the fourth generation; After that, the particles keep the search, but struggle to find better solutions until the eighth generation; Starting from the ninth generation, the particles manage to jump out of the local optima and achieve another round of improvement until the evolutionary process converges at the twelfth generation; once it converges, the fitness of the global best becomes flat until the end. It can be seen that the PSO converges fast due to effort of minimising the search space in the proposed encoding strategy. 

During the stacking and selection step, the evolved block were evaluated by stacking it twice, three times and four times. For the CIFAR-10 dataset, the best classification accuracy was achieved by stacking it three times, which is the final evolved CNN architecture. Furthermore, the final CNN was evaluated by using the fine-tuned SGD optimisation, whose process is illustrated in Fig. \ref{fig:epsocnn_train}. The red line is the training loss, the blue line expresses the error rate on training set, and the green line represents the error rate on test set. From left to right, the sub-figures shows the evaluation processes of 0 to 140 epochs and 140 to 300 epochs, respectively. It can be observed that the evolved CNN was trained fast at the beginning until 40 epochs as shown in the left sub-figure and the reduction of error rate was small after that; when the learning rate was divided at 150 epochs, there was a big plunge in terms of error rate; furthermore, at the send division of the learning rate, a small decrease of the error rate occurred, but it was not significant. Overall, the scheduled learning rate strategy has demonstrated its effectiveness of improving the classification accuracy. 

\subsection{Evolved CNN Architecture}\label{SSS:epsocnn_architecture}

While running the experiment, the server-client infrastructure developed in \cite{wang2019evolving} was adopted, and 10 GPU cards were used, which produced the final CNN architecture in about 9 hours. The hyper-parameters for the evolved block are 23 and 27 for the number of layers and the growth rate, respectively, and the final CNN architecture is composed of three of the evolved block. The computational cost of only spending 9 hours on 10 GPU cards to obtain a good CNN architecture is efficient enough so that the proposed method is feasible and computationally acceptable to be adopted to solve real-life image classification tasks. Taking a step further, the proposed method can be easily adjusted to evolve any of the state-of-the-art CNN blocks such as ResNet blocks. 

\section{Conclusion and Future Work}

To conclude, a PSO method has been proposed to evolve the hyper-parameters of the sate-of-the-art CNN architectures, which is efficient without compromising classification accuracy. The goal has been achieved by minimising the search space with human expertise, learning a transferable block from a small subset of the training set and stacking the learned block multiple times to improve the classification accuracy. By comparing with peer competitors on the CIFAR-10 dataset, the proposed method achieved very competitive performance in terms of the classification accuracy, the model size and the computational cost of searching for the final CNN architecture. The highlight of the proposed method is its efficiency as it outperforms all of the peer competitors, whose results are collected in this paper as to our best effort. In addition, the transferability of the transferable block was evaluated on two other datasets --- CIFAR-100 and SVHN. From the experimental results, the proposed method has achieved promising classification accuracy, especially on CIFAR-100 by outperforming all of the peer competitors. Therefore, the transferable block shows the promising potentiality of achieving good performance in other domains through transfer learning. 

The proposed method has first been evaluated on the CIFAR-10 dataset due to a couple of reasons. The first reason is that CIFAR-10 is good to initially evaluate a neural architecture search method, which usually has a large computational cost, and the second reason is that it is easier to collect the results of peer competitors both from the deep learning community and neural architecture search community. Then, the transferable block learned from CIFAR-10 has been transferred to two other domains --- the CIFAR-100 and SVHN datasets to evaluate the transferability because CIFAR-100 is similar to CIFAR-10 with more classes and SVHN has similar complexity as CIFAR-10 but in a different domain. However, it would be more convincing to evaluate the proposed method on larger datasets such as ImageNet dataset \cite{deng2009imagenet}. Another potential improvement is to explore more state-of-the-art CNN architectures and develop a new method to design better CNN blocks, which consists of multiple types of CNN architecture without compromising the efficiency.

\bibliographystyle{IEEEtran}
\bibliography{epsocnn_cec}

\begin{thebibliography}{10}
\providecommand{\url}[1]{#1}
\csname url@samestyle\endcsname
\providecommand{\newblock}{\relax}
\providecommand{\bibinfo}[2]{#2}
\providecommand{\BIBentrySTDinterwordspacing}{\spaceskip=0pt\relax}
\providecommand{\BIBentryALTinterwordstretchfactor}{4}
\providecommand{\BIBentryALTinterwordspacing}{\spaceskip=\fontdimen2\font plus
\BIBentryALTinterwordstretchfactor\fontdimen3\font minus
  \fontdimen4\font\relax}
\providecommand{\BIBforeignlanguage}[2]{{%
\expandafter\ifx\csname l@#1\endcsname\relax
\typeout{** WARNING: IEEEtran.bst: No hyphenation pattern has been}%
\typeout{** loaded for the language `#1'. Using the pattern for}%
\typeout{** the default language instead.}%
\else
\language=\csname l@#1\endcsname
\fi
#2}}
\providecommand{\BIBdecl}{\relax}
\BIBdecl

\bibitem{krizhevsky2012imagenet}
A.~Krizhevsky, I.~Sutskever, and G.~E. Hinton, ``Imagenet classification with
  deep convolutional neural networks,'' in \emph{Advances in neural information
  processing systems}, 2012, pp. 1097--1105.

\bibitem{simonyan2014very}
K.~Simonyan and A.~Zisserman, ``Very deep convolutional networks for
  large-scale image recognition,'' \emph{arXiv preprint arXiv:1409.1556}, 2014.

\bibitem{he2016deep}
K.~He, X.~Zhang, S.~Ren, and J.~Sun, ``Deep residual learning for image
  recognition,'' in \emph{Proceedings of the IEEE conference on computer vision
  and pattern recognition}, 2016, pp. 770--778.

\bibitem{huang2017densely}
G.~Huang, Z.~Liu, L.~Van Der~Maaten, and K.~Q. Weinberger, ``Densely connected
  convolutional networks,'' in \emph{Proceedings of the IEEE conference on
  computer vision and pattern recognition}, 2017, pp. 4700--4708.

\bibitem{zoph2016neural}
B.~Zoph and Q.~V. Le, ``Neural architecture search with reinforcement
  learning,'' \emph{arXiv preprint arXiv:1611.01578}, 2016.

\bibitem{zoph2018learning}
B.~Zoph, V.~Vasudevan, J.~Shlens, and Q.~V. Le, ``Learning transferable
  architectures for scalable image recognition,'' in \emph{Proceedings of the
  IEEE conference on computer vision and pattern recognition}, 2018, pp.
  8697--8710.

\bibitem{wang2018evolving}
B.~Wang, Y.~Sun, B.~Xue, and M.~Zhang, ``Evolving deep convolutional neural
  networks by variable-length particle swarm optimization for image
  classification,'' in \emph{2018 IEEE Congress on Evolutionary Computation
  (CEC)}.\hskip 1em plus 0.5em minus 0.4em\relax IEEE, 2018, pp. 1--8.

\bibitem{wang2018hybrid}
------, ``A hybrid differential evolution approach to designing deep
  convolutional neural networks for image classification,'' in
  \emph{Australasian Joint Conference on Artificial Intelligence}.\hskip 1em
  plus 0.5em minus 0.4em\relax Springer, 2018, pp. 237--250.

\bibitem{xie2017genetic}
L.~Xie and A.~Yuille, ``Genetic cnn,'' in \emph{Proceedings of the IEEE
  International Conference on Computer Vision}, 2017, pp. 1379--1388.

\bibitem{suganuma2017genetic}
M.~Suganuma, S.~Shirakawa, and T.~Nagao, ``A genetic programming approach to
  designing convolutional neural network architectures,'' in \emph{Proceedings
  of the Genetic and Evolutionary Computation Conference}.\hskip 1em plus 0.5em
  minus 0.4em\relax ACM, 2017, pp. 497--504.

\bibitem{kingma2014adam}
D.~P. Kingma and J.~Ba, ``Adam: A method for stochastic optimization,''
  \emph{arXiv preprint arXiv:1412.6980}, 2014.

\bibitem{sutskever2013importance}
I.~Sutskever, J.~Martens, G.~E. Dahl, and G.~E. Hinton, ``On the importance of
  initialization and momentum in deep learning,'' \emph{ICML (3)}, vol.~28, no.
  1139-1147, p.~5, 2013.

\bibitem{ioffe2015batch}
S.~Ioffe and C.~Szegedy, ``Batch normalization: Accelerating deep network
  training by reducing internal covariate shift,'' \emph{arXiv preprint
  arXiv:1502.03167}, 2015.

\bibitem{nair2010rectified}
V.~Nair and G.~E. Hinton, ``Rectified linear units improve restricted boltzmann
  machines,'' in \emph{Proceedings of the 27th international conference on
  machine learning (ICML-10)}, 2010, pp. 807--814.

\bibitem{eberhart1995new}
R.~Eberhart and J.~Kennedy, ``A new optimizer using particle swarm theory,'' in
  \emph{MHS'95. Proceedings of the Sixth International Symposium on Micro
  Machine and Human Science}.\hskip 1em plus 0.5em minus 0.4em\relax Ieee,
  1995, pp. 39--43.

\bibitem{real2018regularized}
E.~Real, A.~Aggarwal, Y.~Huang, and Q.~V. Le, ``Regularized evolution for image
  classifier architecture search,'' \emph{arXiv preprint arXiv:1802.01548},
  2018.

\bibitem{wang2019hybrid}
B.~Wang, Y.~Sun, B.~Xue, and M.~Zhang, ``A hybrid {GA-PSO} method for evolving
  architecture and short connections of deep convolutional neural networks,''
  \emph{arXiv preprint arXiv:1903.03893}, 2019.

\bibitem{wang2019evolving}
------, ``Evolving deep neural networks by multi-objective particle swarm
  optimization for image classification,'' in \emph{Proceedings of the Genetic
  and Evolutionary Computation Conference}, ser. GECCO '19.\hskip 1em plus
  0.5em minus 0.4em\relax New York, NY, USA: ACM, 2019, pp. 490--498.

\bibitem{krizhevsky2009learning}
A.~Krizhevsky and G.~Hinton, ``Learning multiple layers of features from tiny
  images,'' Citeseer, Tech. Rep., 2009.

\bibitem{cai2018efficient}
H.~Cai, T.~Chen, W.~Zhang, Y.~Yu, and J.~Wang, ``Efficient architecture search
  by network transformation,'' in \emph{Thirty-Second AAAI Conference on
  Artificial Intelligence}, 2018.

\bibitem{elsken2017simple}
T.~Elsken, J.-H. Metzen, and F.~Hutter, ``Simple and efficient architecture
  search for convolutional neural networks,'' \emph{arXiv preprint
  arXiv:1711.04528}, 2017.

\bibitem{liu2017hierarchical}
H.~Liu, K.~Simonyan, O.~Vinyals, C.~Fernando, and K.~Kavukcuoglu,
  ``Hierarchical representations for efficient architecture search,''
  \emph{arXiv preprint arXiv:1711.00436}, 2017.

\bibitem{assunccao2018evolving}
F.~Assun{\c{c}}{\~a}o, N.~Louren{\c{c}}o, P.~Machado, and B.~Ribeiro,
  ``Evolving the topology of large scale deep neural networks,'' in
  \emph{European Conference on Genetic Programming}.\hskip 1em plus 0.5em minus
  0.4em\relax Springer, 2018, pp. 19--34.

\bibitem{miikkulainen2019evolving}
R.~Miikkulainen, J.~Liang, E.~Meyerson, A.~Rawal, D.~Fink, O.~Francon, B.~Raju,
  H.~Shahrzad, A.~Navruzyan, N.~Duffy \emph{et~al.}, ``Evolving deep neural
  networks,'' in \emph{Artificial Intelligence in the Age of Neural Networks
  and Brain Computing}.\hskip 1em plus 0.5em minus 0.4em\relax Elsevier, 2019,
  pp. 293--312.

\bibitem{real2017large}
E.~Real, S.~Moore, A.~Selle, S.~Saxena, Y.~L. Suematsu, J.~Tan, Q.~V. Le, and
  A.~Kurakin, ``Large-scale evolution of image classifiers,'' in
  \emph{Proceedings of the 34th International Conference on Machine
  Learning-Volume 70}.\hskip 1em plus 0.5em minus 0.4em\relax JMLR. org, 2017,
  pp. 2902--2911.

\bibitem{netzer2011reading}
Y.~Netzer, T.~Wang, A.~Coates, A.~Bissacco, B.~Wu, and A.~Y. Ng, ``Reading
  digits in natural images with unsupervised feature learning,'' 2011.

\bibitem{larsson2016fractalnet}
G.~Larsson, M.~Maire, and G.~Shakhnarovich, ``Fractalnet: Ultra-deep neural
  networks without residuals,'' \emph{arXiv preprint arXiv:1605.07648}, 2016.

\bibitem{lee2015deeply}
C.-Y. Lee, S.~Xie, P.~Gallagher, Z.~Zhang, and Z.~Tu, ``Deeply-supervised
  nets,'' in \emph{Artificial intelligence and statistics}, 2015, pp. 562--570.

\bibitem{lin2013network}
M.~Lin, Q.~Chen, and S.~Yan, ``Network in network,'' \emph{arXiv preprint
  arXiv:1312.4400}, 2013.

\bibitem{zagoruyko2016wide}
S.~Zagoruyko and N.~Komodakis, ``Wide residual networks,'' \emph{arXiv preprint
  arXiv:1605.07146}, 2016.

\bibitem{van2006study}
F.~Van~den Bergh and A.~P. Engelbrecht, ``A study of particle swarm
  optimization particle trajectories,'' \emph{Information sciences}, vol. 176,
  no.~8, pp. 937--971, 2006.

\bibitem{he2015delving}
K.~He, X.~Zhang, S.~Ren, and J.~Sun, ``Delving deep into rectifiers: Surpassing
  human-level performance on imagenet classification,'' in \emph{Proceedings of
  the IEEE international conference on computer vision}, 2015, pp. 1026--1034.

\bibitem{huang2016deep}
G.~Huang, Y.~Sun, Z.~Liu, D.~Sedra, and K.~Q. Weinberger, ``Deep networks with
  stochastic depth,'' in \emph{European conference on computer vision}.\hskip
  1em plus 0.5em minus 0.4em\relax Springer, 2016, pp. 646--661.

\bibitem{deng2009imagenet}
J.~Deng, W.~Dong, R.~Socher, L.-J. Li, K.~Li, and L.~Fei-Fei, ``Imagenet: A
  large-scale hierarchical image database,'' in \emph{2009 IEEE conference on
  computer vision and pattern recognition}.\hskip 1em plus 0.5em minus
  0.4em\relax Ieee, 2009, pp. 248--255.

\end{thebibliography}

\end{document}